\documentclass[sigconf]{acmart}
\usepackage{subcaption}
\usepackage{float}
\AtBeginDocument{%
  }

\setcopyright{acmcopyright}
\copyrightyear{2018}
\acmYear{2018}
\acmDOI{XXXXXXX.XXXXXXX}

\copyrightyear{2022}
\acmYear{2022}
\setcopyright{acmcopyright}\acmConference[GeoAI '22]{The 5th ACM
SIGSPATIAL International Workshop on AI for Geographic Knowledge Discovery
(GeoAI '22)}{November 1, 2022}{Seattle, WA, USA}
\acmBooktitle{The 5th ACM SIGSPATIAL International Workshop on AI for
Geographic Knowledge Discovery (GeoAI '22) (GeoAI '22), November 1, 2022,
Seattle, WA, USA}
\acmPrice{15.00}
\acmDOI{10.1145/3557918.3565865}
\acmISBN{978-1-4503-9532-8/22/11}

\begin{document}

\title{highway2vec - representing OpenStreetMap microregions with respect to their road network characteristics}

\author{Kacper Leśniara}
\email{kacper.lesniara@gmail.com}
\affiliation{%
  \institution{Department of Artificial Intelligence,\\ Wrocław University of Science and Technology}
  \city{Wrocław}
  \country{Poland}}

\author{Piotr Szymański}
\email{piotr.szymanski@pwr.edu.pl}
\orcid{0000-0002-7733-3239}
\affiliation{%
  \institution{Department of Artificial Intelligence,\\ Wrocław University of Science and Technology}
  \city{Wrocław}
  \country{Poland}}
\renewcommand{\shortauthors}{Leśniara and Szymański}

\begin{abstract}
Recent years brought advancements in using neural networks for representation learning of various language or visual phenomena. New methods freed data scientists from hand-crafting features for common tasks. Similarly, problems that require considering the spatial variable can benefit from pretrained map region representations instead of manually creating feature tables that one needs to prepare to solve a task. However, very few methods for map area representation exist, especially with respect to road network characteristics. In this paper, we propose a method for generating microregions' embeddings with respect to their road infrastructure characteristics. We base our representations on OpenStreetMap road networks in a selection of cities and use the H3 spatial index to allow reproducible and scalable representation learning. We obtained vector representations that detect how similar map hexagons are in the road networks they contain.
Additionally, we observe that embeddings yield a latent space with meaningful arithmetic operations. Finally, clustering methods allowed us to draft a high-level typology of obtained representations. We are confident that this contribution will aid data scientists working on infrastructure-related prediction tasks with spatial variables. 
\end{abstract}

\ccsdesc[500]{Information systems~Geographic information systems}
\ccsdesc[500]{Information systems~Clustering}
\ccsdesc[500]{Computing methodologies~Learning latent representations}

\keywords{OpenStreetMap embeddings, spatial representation learning, embedding, clustering, road network embeddings}
\begin{teaserfigure}
    \centering
  \includegraphics[width=.6\textwidth]{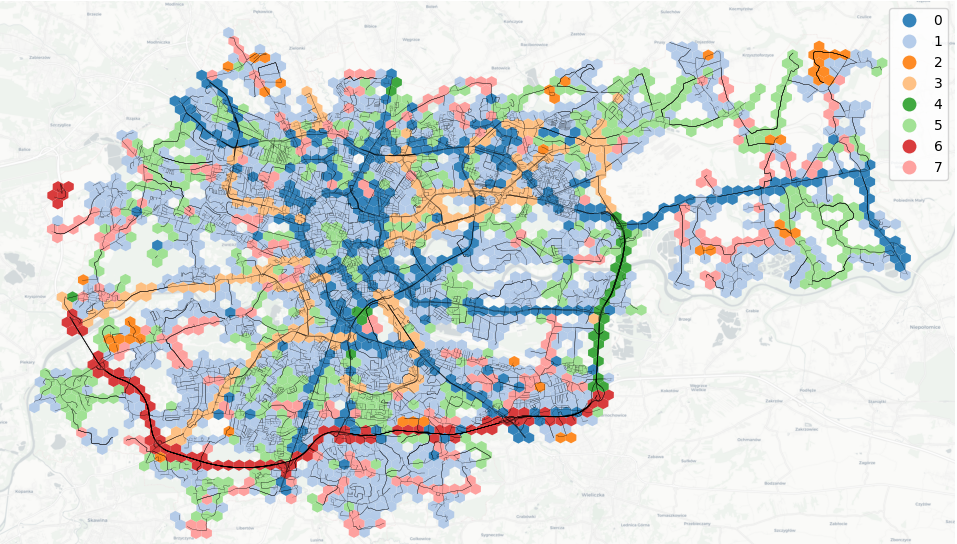}
  \caption{Microregions clustered by the vector representation of their road network characteristics, as discussed in Section~\ref{sec:typology}}
  \label{fig:teaser}
\end{teaserfigure}

\maketitle

\section{Introduction}
Recent years brought advancements in representation learning in almost every field of machine learning. Countless projects no longer have to perform feature crafting, such as counting the number of verbs surrounding a noun or calculating pronunciation speed in audio segments. Deep neural networks and optimization methods brought automatically engineered representations of phenomena happening in large data sets. Map data is no different. We are seeing representation approaches that work on map imagery or certain characteristics of map vector layers. This is possible due to availability of an extensive open source resource - the \textit{OpenStreetMap} \cite{osm}. 

At the same time, we are witnessing a rise in geospatial data, i.e. data that contains a spatial variable related to an underlying process that exhibits a geographical character. The geographical character is often dependent on much more than just the point, line, or polygon which accompanies a certain data point, but also on the - narrower or wider - context, a microregion containing or surrounding a given spatial variable. 

Every city can consist of multiple layers of transportation infrastructure: road networks, railways, airports, and waterways. Solely road infrastructure can consist of multiple layers for different means of transportation. People traveling on the road may choose to walk or ride, including movement by car, bicycle, bus, or motorbike. Each of these ways of transportation can have a dedicated set of tags describing its characteristics in OSM. In this paper, we focus on driveable road networks containing roads accessible always by car. Cars strongly reshaped cities in the last hundred years, and the road infrastructure has become sophisticated. The kind of infrastructure a microregion has follows from local characteristics: land use, historical and current urban planning, and network effects - the role a microregion plays in the infrastructure network of the entire city. 

Methods for microregion representation of map layers exist and are used in the industry. However, as discussed in the background section, they do not yet cover the full spectrum of information present in maps, especially in OpenStreetMap. One of the map data layers that does not have a representation learning method concerning microregions is road networks. The goal of this paper is to provide an approach to learning representations of road networks in OpenStreetMap, which can be further used in predictive tasks. We provide our work as an open source repository on github\footnote{\url{https://github.com/Calychas/highway2vec}}.

The paper is organized as follows: Section \ref{ch:related_works} contains a review of the literature on the road network and microregion representation learning, and Section \ref{ch:data} describes the data sets and pipeline. In Section \ref{ch:proposed_solution} the proposed solution for embedding and obtaining the typology of city regions is presented, followed by Section \ref{ch:experiments}, which contains the results of the experiments and provides the analysis of the obtained groups of microregions. We conclude and discuss future work in Section \ref{ch:conclusion}.

\section{Background on OSM representation learning methods}
\label{ch:related_works}

\subsection{Loc2Vec (2018)}
    \textit{Loc2Vec} \cite{loc2vec} is, to the best of the author's knowledge, the first work that used the representation learning approach to geospatial data. The study aimed to encode locations' surroundings. To achieve that, the authors rasterized the region around a given point and passed that to the deep convolutional neural network, which generated semantic embedding. Instead of treating a single tile as a three-channel RGB image, the authors used a 12-channel tensor, where each channel contained a different type of information (road network, land cover, amenities, etc.). The data was gathered from \textit{OpenStreetMap}. A self-supervised method was used to transform rasters into embeddings - triplet networks with a triplet loss. It is based on the concept of positive and negative instances for a particular instance, where the former describes instances that are close to the main one and the latter these instances that are far. The authors explored the obtained embedding space thoroughly using dimensionality reduction techniques, random walking, and performing interpolation and calculation on the embeddings.

\subsection{Zone2Vec (2018)}
    One of the works that generated embeddings of regions was \textit{Zone2Vec} \cite{zone2vec}. The zones were created by segmenting the street network of a city by its major roads. To generate the zone's embedding that incorporates its characteristics, the authors considered the relationships between the zones using trajectories and intrinsic properties of such zones. Each trajectory was projected onto the urban zones, which transformed the trajectories' locations into a zone sequence. In their study, the authors used Beijing Road Network and GPS trajectories generated by taxis. Additionally, point of interest (POI) data was used that contained longitude, latitude, and category. To enrich the data further, the authors chose to add social media data (from Sina Weibo) that included check-in locations in Beijing. Embeddings are created by employing a multi-label classification method on POI categories contained within a zone with the help of the \textit{Skip-gram} \cite{skip-gram} model. It is followed by a spectral clustering approach on the generated embeddings to determine similar zones. 

\subsection{ZE-MOB (2018)}
    \textit{ZE-MOB} \cite{ze-mob} is another approach to generating meaningful zone embeddings in terms of urban functions. The authors proposed a framework in which taxi trajectories are utilized. Human mobility patterns were extracted from the real-world urban dataset of New York City, which contained origin-destination pairs (mapped to specific zones), time of departure and arrival, as well as travel distance. The US Census Bureau designed the zones used in this work, they are separated by major roads, and the areas are small enough to support meaningful analyses.

\subsection{RegionEncoder (2019)}
    One of the complete solutions to multimodal region embedding is \textit{RegionEncoder} proposed in \cite{regionencoder}. The authors define an end-to-end framework enabling, as the authors describe as, \textit{Learning an Embedding Space for Regions} (\textit{LESR}). In this paper, the types of data used are as follows: cities are grid partitioned and treated as a spatial graph, taxi mobility data model inter-region flow and weight edges, POI category is used to describe the functional distribution for each region, and satellite imagery is used to obtain the ground RGB image. To merge these multiple modalities, the authors used three methods. Denoising Autoencoder was used to embed satellite images. For the other types of data \textit{Graph Convolutional Network} (\textit{GCN}) \cite{kipf2017semisupervised} was applied. The output of both these methods was concatenated and treated as an input to the \textit{Multilayer Perceptron} (\textit{MLP}), to unify two distinct embeddings into one. The third method acts as a discriminator, which learns to predict if the embeddings belong to the same region. The final latent space is obtained by getting the last hidden state of the discriminator. As the authors state, all models were trained jointly.
\subsection{Modelling urban networks using Variational Autoencoders (2019)}
    In \cite{Kempinska2019ModellingUN} a method for embedding regions based on the road network is proposed. This work uses \textit{Variational Autoencoders} (\textit{VAE}) with convolutional layers to create a latent space based on street networks from \textit{OpenStreetMap}. The road infrastructure is treated as grid partitioned images (each cell 3 x 3 km). Each image's pixel is represented by $0$ if there is no road and $1$ if there is. The authors obtained the embedding space that can capture street network density, structure, and orientation.

\subsection{HRNR (2020)}
    Another work that tries to capture the graph structure of the road infrastructure is \textit{Hierarchical Road Network Representation} (\textit{HRNR}) \cite{hrnr}. The approach is constructing the three-level architecture that tries to model two probability distributions. One such distribution would be the road segment to region assignment and the other region to zone assignment. Road segments incorporate some metadata (position, segment type, length). In modelling structural regions Spectral Clustering is used with the addition of \textit{Graph Attention Network} (\textit{GAT}) \cite{gat}. Functional zones also utilize \textit{GAT}. Trajectory data is used to capture functional characteristics. The first two steps of the hierarchical update are performed with the use of \textit{GCN}. Firstly, the zone-level update is done to update the zone embeddings, followed by the region-level update. At the end (with the use of \textit{GAT}), the segment-level update is performed. 

\subsection{RLRN (2020)}
    The authors of \cite{rn2vec} designed a framework called \textit{Representation Learning for Road Networks} (\textit{RLRN}) in which the neural network model \textit{Road Network to Vector} (\textit{RN2Vec}) was proposed. The framework consists of two phases. In Phase I, the authors argue with \textit{Node2vec}'s \cite{node2vec} approach to sampling walks, writing: 'random walk-based sampling is not suitable for road networks, because mobile road users do not move randomly. The authors' solution is to generate the data by sampling the shortest paths and incorporating paths generated by users. Phase II neural network model is proposed, which is actually compounded of three submodules: \textit{Intersection of Road Network to Vector} (\textit{IRN2Vec}) (introduced by the authors in \cite{irn2vec}) focused on intersections, \textit{Segment of Road Network to Vector} (\textit{SRN2Vec}) taking into account road segments, and the final \textit{Intersection and Segment of Road Network to Vector} (\textit{ISRN2Vec}) incorporating the embeddings generated by the previous two models. The framework was evaluated on five different downstream tasks and three cities, and the results obtained were superior to the baseline models.

\subsection{Hex2Vec (2021)}
    The first work that divided cities using spatial indexes is \textit{Hex2Vec} \cite{hex2vec}. The authors partitioned the space using \textit{Hexagonal Hierarchical Spatial Index} (\textit{H3}) \cite{h3}, and for each of the hexagons, the embedding was obtained. The latent space representations were created by incorporating the geospatial knowledge of POIs and their tags. The data was gathered fully from OpenStreetMap. The proposed method is based on \textit{Skip-gram} \cite{skip-gram} model with negative sampling. The neighbors of a particular hexagon are treated as positive samples, and the rest (with a margin of one from the neighbors) are treated as negative instances. As the authors argue, such an approach is possible due to Tobler's law \cite{first_law_of_geography} (as also argued in \cite{loc2vec}).

\section{Data}
\label{ch:data}
The mentioned literature in the field provides knowledge of how this task can be achieved. The assumption is to create an unsupervised method which provides portability between the cities. Data required to generate embeddings for different zones must be easily accessible to the general public, and the proposed method must not use additional proprietary data sources. To this date, no studies have explored road networks numerically (in the sense of incorporating a greater number of features) in the context of spatial indexes. Recently, however, such spatial indexes gained considerable interest, and this thesis explores this topic even further. All related works performed similarity detection on the obtained latent space to some extent. To create a universal and portable method across multiple cities, we employ \textit{OpenStreetMap} (\textit{OSM}) \cite{osm}, which is an \textit{open data} source. \textit{OSM} is a map data service that is built by the community. The components of the \textit{OSM}'s data model are as follows:
    
    \begin{itemize}
        \item nodes - particular points on the globe, defined by latitude and longitude,
        \item ways - linear features and areas,
        \item relations - the structure that describes the relationship of other components (including other relations).
    \end{itemize}
    
\begin{figure}[!htp]
        \centering
        \includegraphics[width=\linewidth]{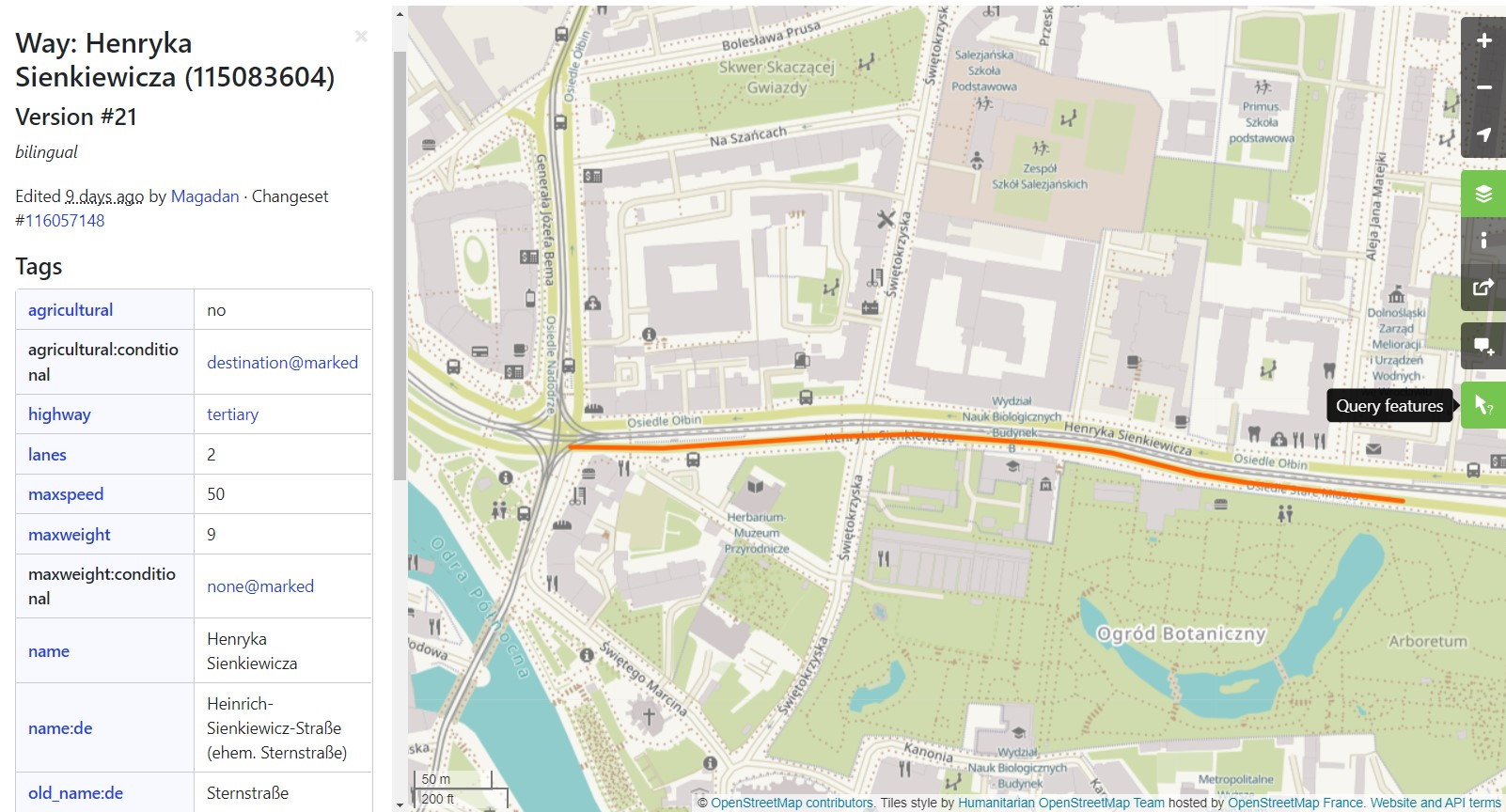}
        \caption[A road segment returned by the \textit{OSM} query in Wrocław, Poland]{A road segment returned by the \textit{OSM} query in Wrocław, Poland. The left sidebar contains the meta-information (tags) about the way}
        \label{fig:osm_query}
    \end{figure}

    The road infrastructure data contains roads as ways and intersections as nodes. One can query the database indirectly using the \textit{Overpass API} on the \textit{OSM} website. Figure \ref{fig:osm_query} shows the results of a query focused on obtaining road segments.
    
\begin{figure}[!htp]
        \centering
        \includegraphics[width=\linewidth]{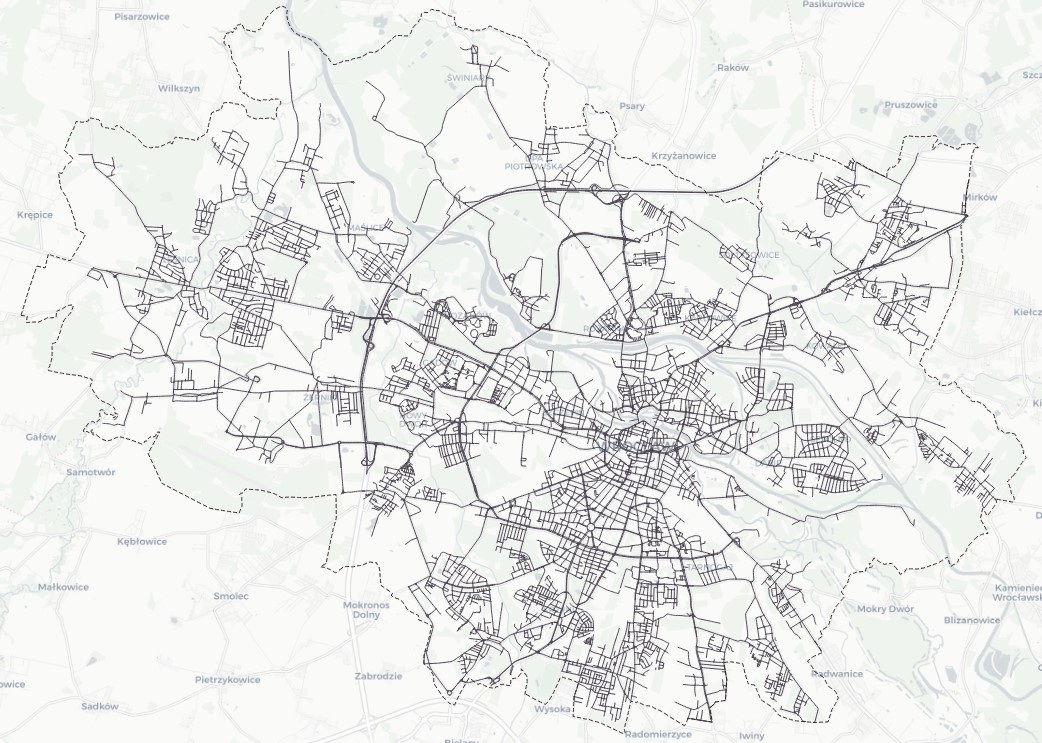}
        \caption{Road network}
        \label{fig:wroclaw_edges}
        \caption{Obtained driveable road network for 'Wrocław, Poland' query}
    \end{figure}

    To download the data automatically for a larger number of cities, one can use the \textit{Overpass API} directly. However, the structure of the returned results is far from optimal. The \textit{way} elements are divided into segments, and such division in the analysis context is unnecessary. We use the \textit{OSMnx} library \cite{osmnx} to obtain road networks for multiple global cities. The library simplifies the road network and returns a graph with a proper division between intersections and roads. It also returns the features associated with the objects. The driveable road network obtained for the given polygon is depicted in Figure \ref{fig:wroclaw_edges}.
    
\subsection{Cities}
\begin{figure}[!htp]
        \centering
        \includegraphics[width=\linewidth]{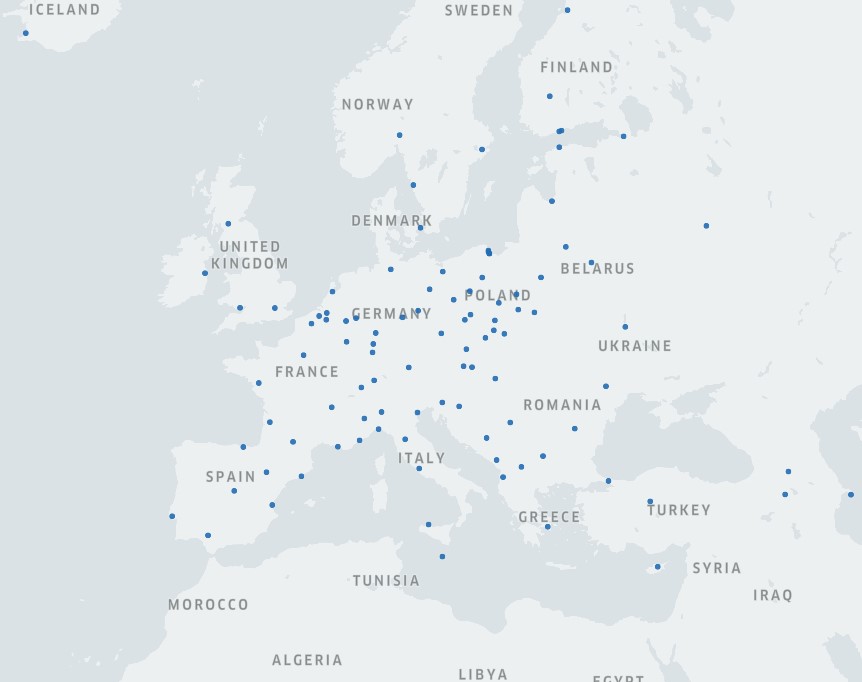}
        \caption{Downloaded cities in Europe}
        \label{fig:cities_europe}
    \end{figure}

    As we concentrate mainly on urban areas, we assess the tags available in the \textit{OSM} on a larger number of cities across the globe were downloaded than are actually used in the proposed solution. We downloaded $117$ cities on \textit{2021-12-18} using the \textit{OSMnx} library, which are shown in Figure \ref{fig:cities_europe}.
    
\begin{figure}[!htp]
        \centering
        \includegraphics[width=\linewidth]{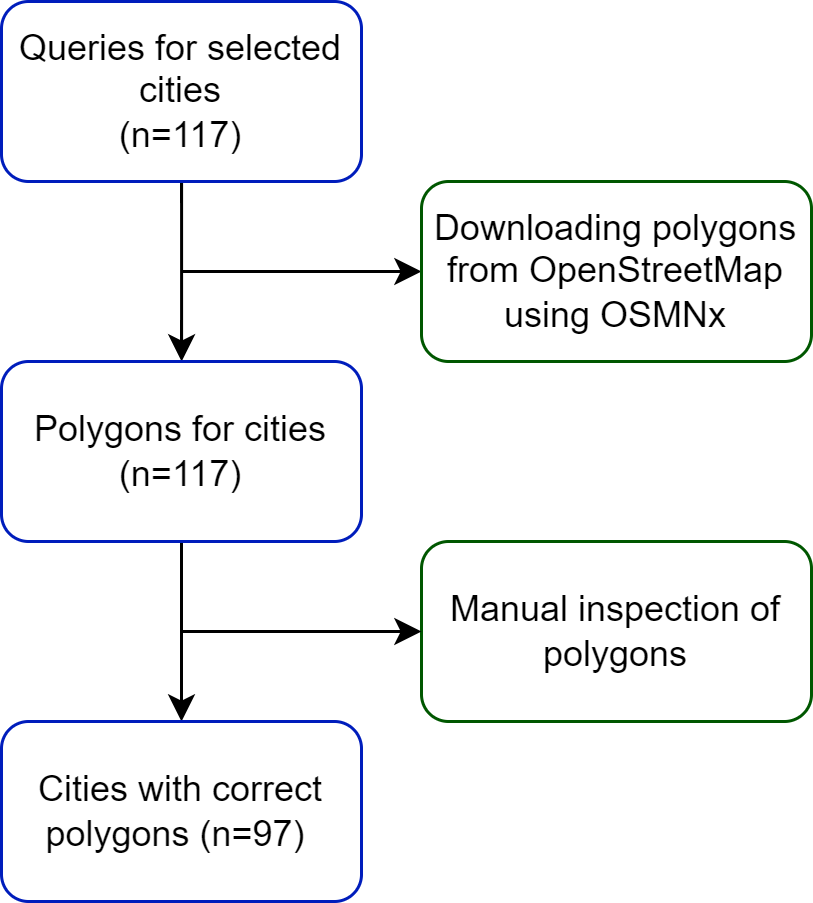}
        \caption{City selection process}
        \label{fig:city_selection}
    \end{figure}

    The process of downloading the data uses geocoding, which converts the string query into geographic coordinates, and in the end, returns a polygon of a city. This process is visualized in Figure \ref{fig:city_selection}. 
    
    Every city was manually checked to validate the quality of a query, and some cities were dropped from further analysis due to an incorrect polygon. The final distribution of $97$ selected cities.
    
\subsection{Features}
    Every object in the \textit{OSM} can contain multiple tags in the form of key-value pairs. As the \textit{OSM} is a community-driven service with no strict enforcement on the data added (only guidelines exist), various keys and values may exist. It allows the service to be flexible, but at the same time, it lowers the quality of the data. \textit{Taginfo} \cite{taginfo} service was used to discover the available values of the features and their frequencies. It is a website that aggregates the tags and shows statistics about the values and tags used by the community. In addition, \textit{OSM Wiki} \cite{osm_wiki} was utilized to understand the tags and possible values for the tags. Furthermore, a manual inspection of the data downloaded for $97$ selected cities across the globe was conducted. Based on these three approaches, the following keys were taken into consideration for the road segments of driveable road networks:
    \begin{itemize}
        \item \textbf{oneway} - whether the road is one-way,
        \item \textbf{highway} - rank of the road,
        \item \textbf{surface} - physical surface, structure, composition,
        \item \textbf{maxspeed} - maximum legal speed limit,
        \item \textbf{lanes} - number of traffic lanes,
        \item \textbf{bridge} - type of bridge that the way is on,
        \item \textbf{junction} - type of junction that the way forms itself,
        \item \textbf{access} - restrictions on the use,
        \item \textbf{tunnel} - type of an underground passage,
        \item \textbf{width} - actual width of a way.
    \end{itemize}

\begin{figure}[!htp]
        \centering
        \includegraphics[width=\linewidth]{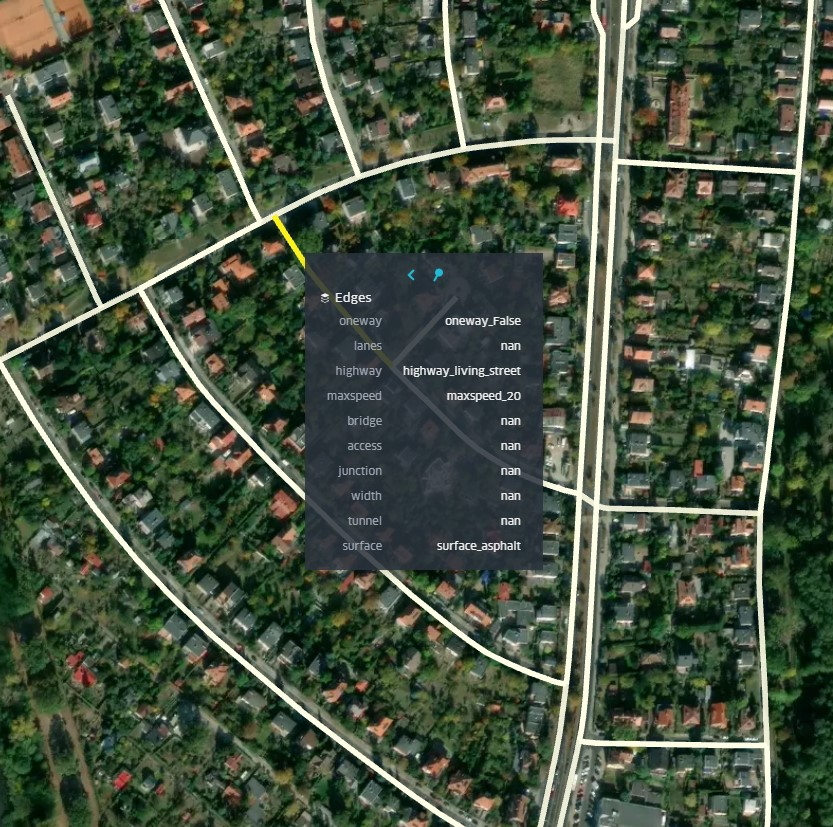}
        \caption{An example of features for one road segment}
        \label{fig:road_example}
    \end{figure}

\begin{figure}[!htp]
        \centering
        \includegraphics[width=\linewidth]{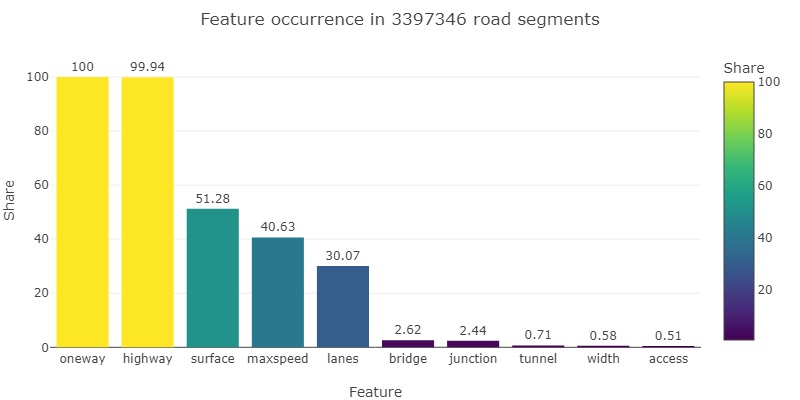}
        \caption{Feature keys occurrences in the obtained dataset of 97 selected cities}
        \label{fig:feature_shares}
    \end{figure}

    Furthermore, every key has many values that need to be extracted, consolidated (e.g., brought to the same unit, rounded), and mapped. Then, the final set of features can be selected that will be treated as an input to the model. If the road segment contains a particular tag, the value is $1$; otherwise $0$. An example of features for one road segment is visible in Figure \ref{fig:road_example}. Feature coverage varies between the cities. Figure \ref{fig:feature_shares} shows the key occurrence in the gathered data. 
    
    One can see that the most occurring features are \textbf{oneway} and \textbf{highway}, which are the essential information on road networks. No navigation system could use the road infrastructure data with these tags missing. Around 62\% of roads are not one-way roads, and unsurprisingly, because the data is gathered solely from cities, the most common \textbf{highway} value is \textbf{residential} (around 58\%). 
    
    The next common tags, that also relate to all of the roads, are \textbf{surface} (51\%), \textbf{maxspeed} (40\%) and \textbf{lanes} (30\%). As we focus more on the highly urbanized areas in Europe, almost all of the streets are either \textbf{asphalt} or \textbf{paved}. The most common \textit{maxspeed} values are $50$ and $30$, followed by $60$, which is also highly correlated to picking cities as the source of the data. Small \textbf{lane} values ($1$-$4$) are also popular in city road infrastructure.
    
    A lot less roads contain \textbf{bridge}, \textbf{junction}, \textbf{tunnel}, or \textbf{access} tags, but these tags can be an important factor in differentiating the microregions that the roads belong to. The last tag, however not widely adopted, is \textbf{width}. It gives a better understanding of the physical form of the road. The most common widths are 5m and 6m, which for cities is highly correlated with the number of lanes, which is $2$. As an example, in Poland, the minimal width for a lane of a residential road is between 2,5m to 3,5m \footnote{\url{http://isap.sejm.gov.pl/isap.nsf/DocDetails.xsp?id=WDU20160000124}}.

\subsection{Microregions} \label{sec:ch3_microregions}
    \textit{H3} spatial index approach to partitioning cities into microregions  was chosen. Based on the manual inspection of regions, the resolution value of 9 is believed to give the best trade-off between generalization and noise. Too tiny regions would mostly be able to capture single road segments, and as a consequence, the region embedding would be the same as a single road embedding. Too large regions, however, would contain roads of many different types and characteristics, leading to little regional variation. The H3 spatial index was chosen as separating the space into hexagons allows regions to an edge with all their neighbors, and the distance between the center of each region and the centers of each of its neighbors is the same.

    \begin{figure}[!htp]
        \centering
        \includegraphics[width=\linewidth]{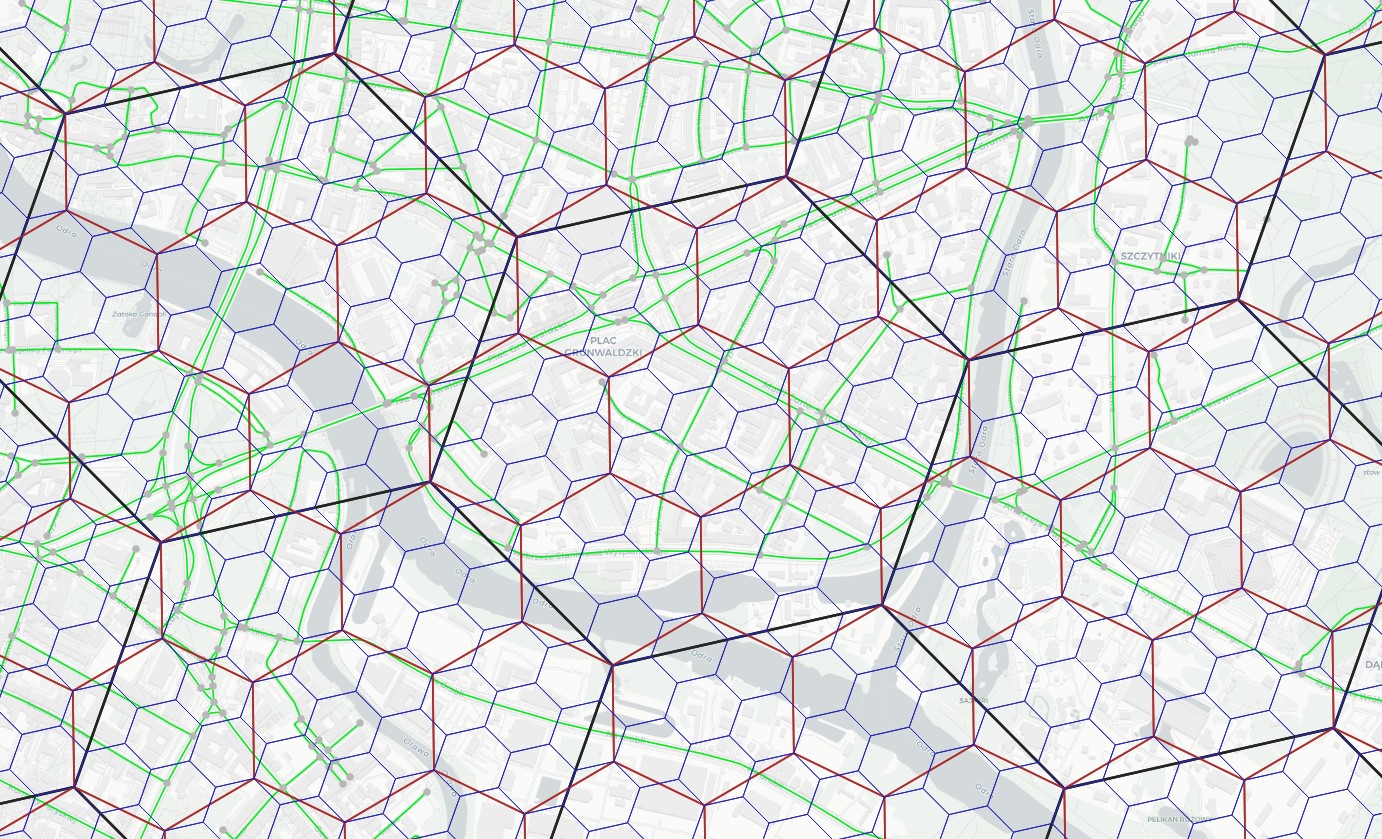}
        \caption[Hex resolution comparison for the Grunwald Square and surroundings in Wrocław, Poland]{Hex resolution comparison for the Grunwald Square and surroundings in Wrocław, Poland. The hexagon color reflects the resolution: black - 8, red - 9, blue - 10. Roads and intersections are shown in green and grey, respectively}
        \label{fig:hex_resolution_comparison}
    \end{figure}

\section{Proposed solution}
\label{ch:proposed_solution}
\begin{figure}[!htp]
        \centering
        \includegraphics[width=\linewidth]{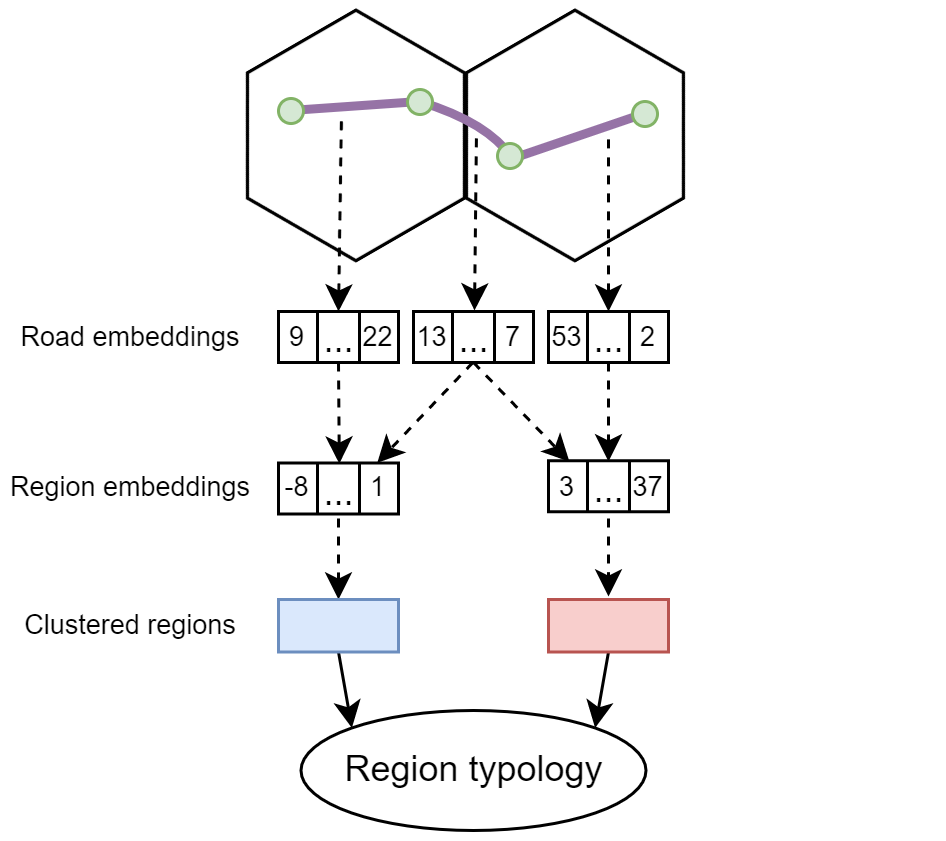}
        \caption{Methodology of the proposed solution}
        \label{fig:method_framework}
    \end{figure}

In this section, a method for obtaining vector representations of region-aggregated road infrastructure information is presented. It consists of three phases: partitioning the area into regions, feature aggregation, and representation learning. Finally, it is explained how the obtained vectors will be used to cluster regions into interpretable road infrastructure profiles. The methodology of the proposed solution is depicted in Figure \ref{fig:method_framework}.

\subsection{Region embedding}
    In this section, the approach to generating region embeddings will be introduced. Firstly, the process of obtaining the representation of a particular road segment is presented, then the transformation from a road embedding space to a microregion latent space is described. As discussed in Section \ref{sec:ch3_microregions}, \textit{H3} \cite{h3} spatial index with resolution $9$ was used to partition the urban area into microregions. The utilization of a universal hierarchical spatial index provides portability across different cities and studies. It also enables combining the obtained representations for particular microregions. One can merge the embeddings to achieve more robust representations.
    
    \subsubsection{Autoencoder}
        \label{subsec:autoencoder}
        To generate embeddings for road segments that are described by the features, the \textit{Autoencoder} \cite{autoencoder} neural network model was used. It is an unsupervised approach, at the core, high-dimensional unlabeled data need to be represented in a low-dimensional space, without the need of labeling that data.
        \begin{figure}[!htp]
            \centering
            \includegraphics[width=\linewidth]{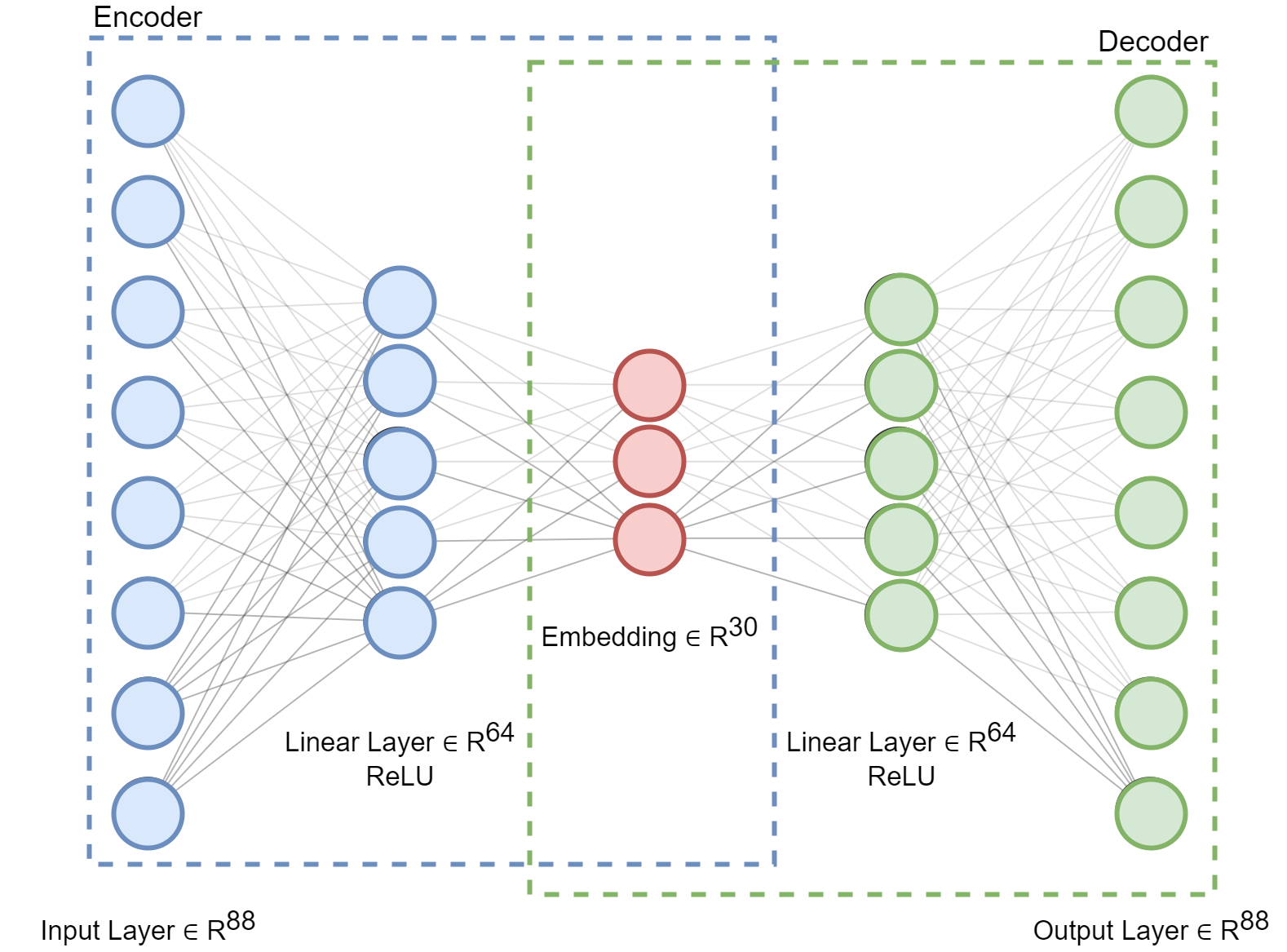}
            \caption{Architecture of the used \textit{Autoencoder} method}
            \label{fig:autoencoder}
        \end{figure}
        
        The \textit{Autoencoder} consists of two parts - an \textit{encoder} and a \textit{decoder}. The former's responsibility is to reduce the dimensionality of the input data $X$, and in the end, it outputs the lower-dimensionality version - an embedding $h$. The latter is to reconstruct the input $X'$ from the obtained embedding. The architecture of the \textit{Autoencoder} proposed in this paper is depicted in \ref{fig:autoencoder} and also described below.
        \paragraph{Encoder:}
        \begin{itemize}
            \item Input layer - 88 features,
            \item Linear layer with ReLU - 64 dimensions,
            \item Embedding layer - 30 dimensions.
        \end{itemize}
        \paragraph{Decoder:}
        \begin{itemize}
            \item Embedding layer - 30 dimensions,
            \item Linear layer with ReLU - 64 dimensions,
            \item Linear layer - 88 dimensions.
        \end{itemize}
        
        The \textit{Autoencoder} uses a backpropagation algorithm \cite{backprop}. The training is based on minimizing the reconstruction loss. There are several loss functions suited for different tasks: \textit{Mean Squared Error} (\textit{MSE}) or \textit{Binary Cross Entropy} (\textit{BCE}). Using \textit{BCE} would be a natural approach to handling binary features of the road segments, however, the obtained embeddings proved to be sub-par to the ones generated with \textit{MSE}. In the end, the loss was chosen to be the \textit{Mean Squared Error}, which for the input X is defined as follows:
        \begin{equation}
            L  = MSE = \frac{1}{N} \sum_{n=1}^N (X_{n} - X'_{n}).
            \label{eq:mse}
        \end{equation}
        
    \subsubsection{Roads aggregation}
\begin{figure}[!htp]
            \centering
            \includegraphics[width=\linewidth]{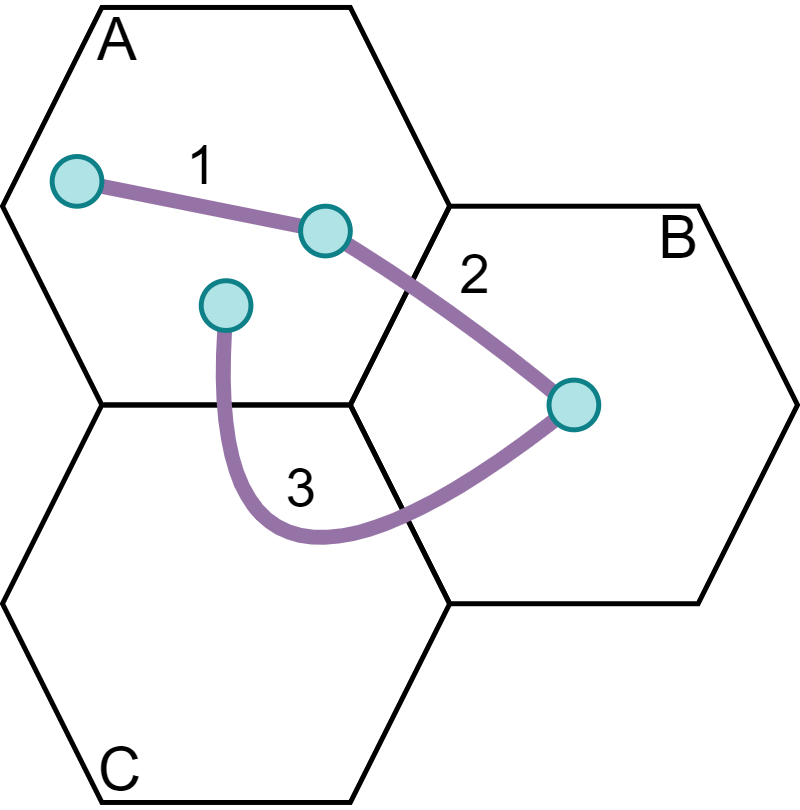}
            \caption[Road features aggregation into hexagons]{Road features aggregation into hexagons. Road $1$ is aggregated to hexagon $A$, road $2$ is aggregated to hexagons $A$ and $B$, road $3$ is aggregated to hexagons $A$, $B$, and $C$}
            \label{fig:feature_aggregation}
        \end{figure}

        To generate the region representations, the road segments within must be aggregated. All roads that intersect with a hexagon, count to a particular microregion. A visual representation for different cases of aggregation is shown in Figure \ref{fig:feature_aggregation}.
    
        One can choose the aggregation function of the road embeddings. The embeddings that belong to a particular microregion are \textit{averaged}, which results in one representation for the region in the same latent space as the road segment embeddings. As the co-occurrence of the features in the road segments is essential, aggregation cannot be performed before the road segment embedding process. Simply counting the tags that a region has, followed by embedding that region, will lose information about the characteristics of particular roads. The final profile of the region would not represent the contained roads accurately.
        
\subsection{Similarity detection}
    Having the urban space partitioned into microregions, and the representations in the latent space of such regions, one can detect similarities between them. Using the knowledge gathered during the review of the literature in Section \ref{ch:related_works} multiple approaches to visualizing and defining the obtained latent space will be presented. 
    
    \subsubsection{Agglomerative Clustering}
        One of the approaches to detect similarities between regions is to find the groups that the underlying data naturally forms. Based on that, the analyses can be done for particular groups instead of single instances, which can provide a more insightful comparison. In addition, obtaining the hierarchy of groups may further aid the process of determining the region typology. For that purpose, the \textit{hierarchical clustering} method was used - \textit{Agglomerative Clustering} \cite{hierarchical_clustering}. It is based on the bottom-up approach, where at the beginning each instance is in its own cluster, and the clusters are merged successively. The pairs which minimize the linkage criterion (difference between the sets) are merged. The \textit{Ward} \cite{ward} linkage criterion was used, which focuses on minimizing the variance of the groups. To compute the linkage, a distance metric is needed, and for that, the \textit{euclidean} distance was used. Looking at the feature distribution in the clusters, one can identify the characteristics of particular groups. Considering also multiple levels of hierarchical groups, a region typology based on road infrastructure in cities will be established.
        
    \subsubsection{Latent space analysis}
        To analyze the obtained latent space, dimensionality reduction techniques will be used to provide embeddings suited for visualizations. \textit{Principal Component Analysis} (\textit{PCA}) \cite{PCA} will aid in showing the diversification of regions in cities. It will lower down the dimensionality of the embeddings to $3$, and later convert the obtained values to the \textit{RGB} color space. Furthermore, \textit{t-Distributed Stochastic Neighbor Embedding} (\textit{t-SNE}) \cite{tsne} will enable looking into the embedding space in $2$ dimensions, to see how the space is laid out. Additionally, operations on the obtained embeddings will be performed (similar to \cite{operations_in_embedding_space}) - adding and subtracting - to find if the semantic meaning of the real-world objects is preserved.

\section{Experiments}
\label{ch:experiments}
In this section, we present the experimental results. First, the setting for the model training will be described. Then the hierarchical clustering results will be shown, followed by the defined region typology. Finally, the visualizations of the obtained latent space will be presented and the operations will be performed.

\subsection{Setting}
    Training and testing the \textit{Autoencoder} neural network proposed in Subsection \ref{subsec:autoencoder} was performed on a subset of data described in Section \ref{ch:data}. The data consists solely of $13$ Polish cities. It contains $191868$ road segments and $22991$ \textit{H3} microregions, that contain at least one road. Other hexagons are not considered. Similarity detection and latent space visualizations will also be performed on the same dataset, however, the clustering results will only be shown for $6$ selected cities.
    
    The network architecture was defined in Subsection \ref{subsec:autoencoder}. The proposed pipeline, however, needs additional hyperparameters defined. The values are as follows:
    \begin{itemize}
        \item optimizer - \textit{Adam} with learning rate $0.001$,
        \item batch size - $200$,
        \item epochs - $50$,
        \item test dataset ratio - $0.2$,
        \item t-SNE perplexity - $100$.
    \end{itemize}

\subsection{Similarity detection}
    This section will provide the results for multiple levels of hierarchical clustering. At the end, the region typology will be proposed.
    
    \subsubsection{Hierarchical clustering}
\begin{figure}[!htp]
            \centering
            \includegraphics[width=\linewidth]{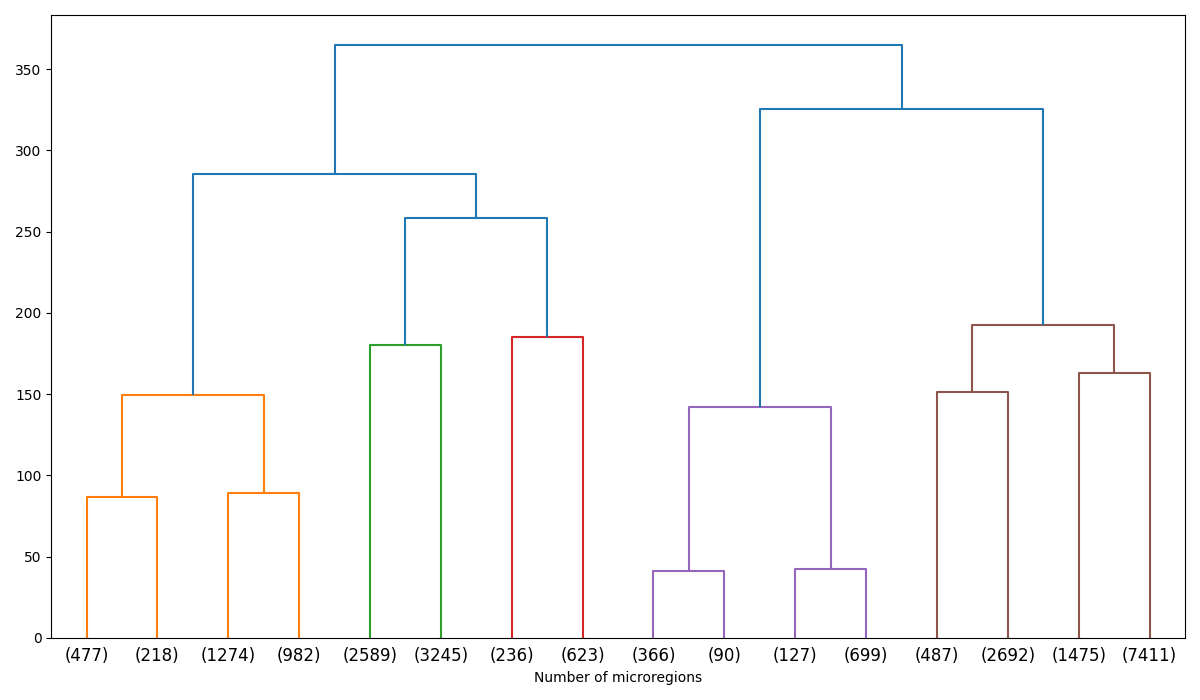}
            \caption{Dendrogram for the dataset}
            \label{fig:dendrogram}
        \end{figure}

        \textit{Hierarchical clustering} provides a cluster tree that represents the successive splits of groups. Such visualization is called \textit{dendrogram} and is shown in Figure \ref{fig:dendrogram}. It can give an overview of how the partitioning process will look like. As stated in Section \ref{ch:proposed_solution}, the \textit{Agglomerative clustering} method was used in generating the data splits.
    
        The rest of this section will be organized as follows - for every division a description will be provided, accompanied with visualizations of clustered microregions for selected Polish cities. Additionally, a difference between the two divided clusters will be presented. A cluster is called \textit{new} if the overall number of microregions in this cluster is fewer than in the other \textit{old} cluster. The difference for a particular tag between the \textit{new} and \textit{old} groups is simply the difference in the share of this tag in the corresponding feature.
        
    \subsubsection{Region typology}
    \label{sec:typology}
        This section summarizes the results of the \textit{hierarchical clustering} experiments carried out on 13 Polish cities. The groupings of the microregions for different splits were discussed and based on these results, the region typology will be proposed.
        The final number of clusters was determined to be $8$, as further splits do not provide meaningful information and oftentimes are a product of artifacts in the data (e.g., incomplete tag information for road segments). The obtained embeddings are capable of generating a diversified profile of microregions across multiple cities. The grouped zones for selected urban areas are shown in Figure \ref{fig:exp_hex_8}, and the final profiles of microregions can be described as follows:
        \begin{itemize}
            \item $0$ - high-capacity regions containing arterial roads,
            \item $1$ - residential, paved regions with good quality of road infrastructure,
            \item $2$ - residential, unpaved regions with low quality of road infrastructure,
            \item $3$ - regions complementing main road network,
            \item $4$ - high-capacity, high-speed regions and bypasses, 
            \item $5$ - estate regions and connectors, 
            \item $6$ - motorways,
            \item $7$ - traffic collectors and connectors.
        \end{itemize}
        
        The obtained region typology properly diversifies the city regions into multiple functionally meaningful groups in terms of road infrastructure. The final share of features across all clusters is depicted in Figure \ref{fig:features_clustering}. The proposed solution generates satisfactory results and enables automatic discovery of the city structure. Furthermore, it validates the quality of the obtained latent space and the used methodology.

\begin{figure}[!htp]
        \centering
        \includegraphics[width=\linewidth]{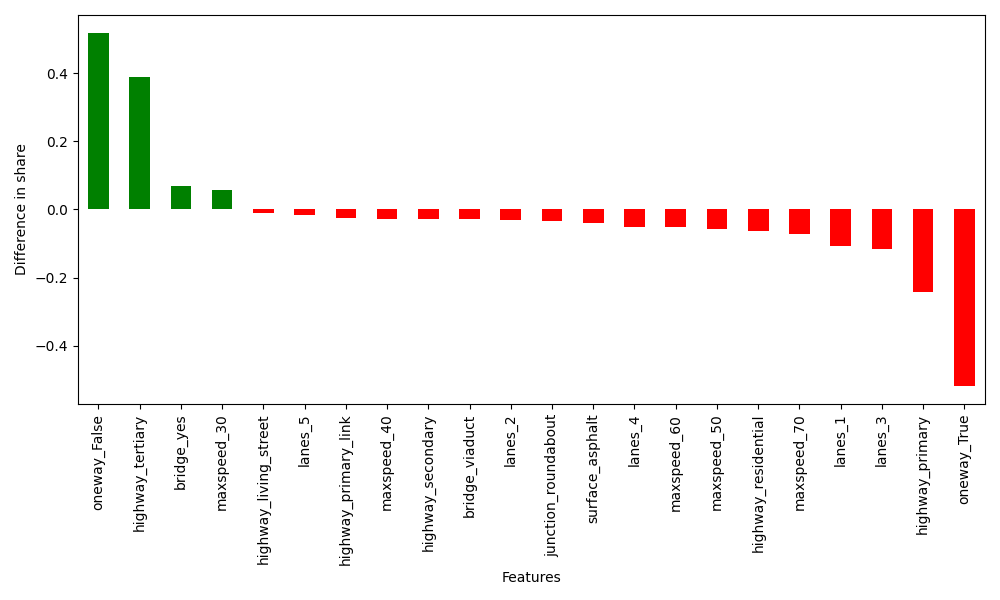}
        \caption{Cluster difference between $7$ and $0$ for $8$ clusters}
        \label{fig:cluster_difference_8}
    \end{figure}

    The \textbf{blue cluster 0} regional profile was still unclear and contained a lot of different types of main roads in cities. The $8$ cluster split happened exactly on that one. Let's start with the Figure \ref{fig:cluster_difference_8}. The new \textbf{light red cluster 7} regions are made up of tertiary two-way roads, and as a consequence, the other cluster contains primary one-way road infrastructure. As a result, the \textbf{light red cluster 7} profile is mostly described by collector roads in the city. Comparing it to the maps (Figure \ref{fig:exp_hex_8}), one can see that these regions' purpose is to connect areas and provide an infrastructure responsible for accessing the main arterial roads within a city. Take \textit{Wrocław} for an instance. The entire right side of a city depends on the new cluster traffic-wise, as well as on \textbf{light orange cluster 3}. \textbf{Light red cluster 7} regions grant residential areas (\textbf{light blue}) access to the main part of the road network. Further splits introduce noise, and meaningful diversification of the microregions becomes difficult. Often, the partitioning results from artifacts in the data (e.g., missing tags).
    
\subsection{Latent space analysis}
\begin{figure}[!htp]
        \centering
        \includegraphics[width=\linewidth]{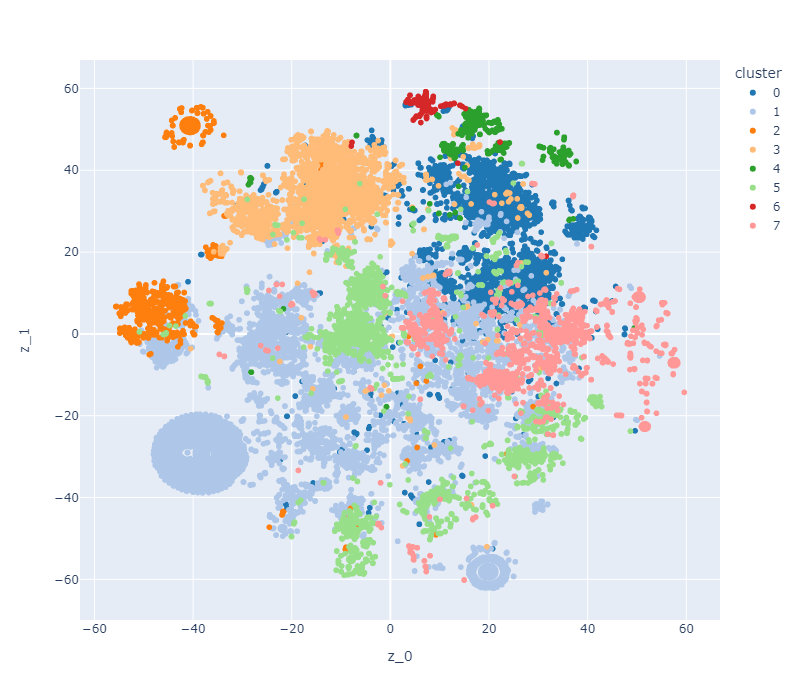}
        \caption{Latent space visualization using \textit{t-SNE} and colour-coded in $8$ clusters}
        \label{fig:latent_space_visualization}
    \end{figure}

    To further validate the quality of the representations, the obtained space can be analyzed. As a reminder, the road segments containing \textit{OSM} tags were embedded and then aggregated (\textit{averaged}) to their corresponding microregions. The resulting regional vectors are in the same space as the road embeddings. \textit{t-SNE} was used to visualize the microregion representations in $2$ dimensions. Additionally, the grouping from the previous section was utilized. The result is depicted in Figure \ref{fig:latent_space_visualization}. 
    
    One can see that the clusters are rather separated. \textbf{Green cluster 4} and \textbf{red cluster 6} are both represented by high-capacity, high-speed roads, but the distinction between motorways and trunk roads is apparent. \textbf{Light orange cluster 3} regions are also separated from others as the group focuses mostly on secondary roads. The big initial \textbf{blue cluster 0} got divided to the point that eventual further splits could contain a lot of noise. Moving to the second initial \textbf{light blue cluster 1}, which contained paved residential areas - it did not separate that well. Two large concentrations of regions are visible around $(20, -60)$ and $(-40, 30)$. These could be separated in further splits. The rest of this cluster is spread, and that could mean that the functional profile of this group can be fuzzy. The \textbf{orange cluster 2} also contains residential roads but focuses on the unpaved aspect of the road's surface. The \textbf{light green cluster 5} has one highly concentrated zone and four other smaller zones. Probably the difference is in estate areas and connectors, and the division would happen in further splits. The last cluster, \textbf{light red cluster 7}, containing mainly distributor roads is situated near \textbf{blue cluster 0}, which contains arterial roads. It makes sense in terms of the functional purposes of the microregions.
           
    \subsubsection{Semantic arithmetics}
        The last experiment will perform arithmetic operations on the obtained embedding space to validate if the semantic meaning of the real-world objects is preserved. We show how \textbf{addition} between two embedded microregions exhibits semantic phenomena.
        
            \begin{figure}[!h]
                \centering
                \begin{subfigure}[t]{0.32\textwidth}
                    \centering
                    \includegraphics[width=\linewidth]{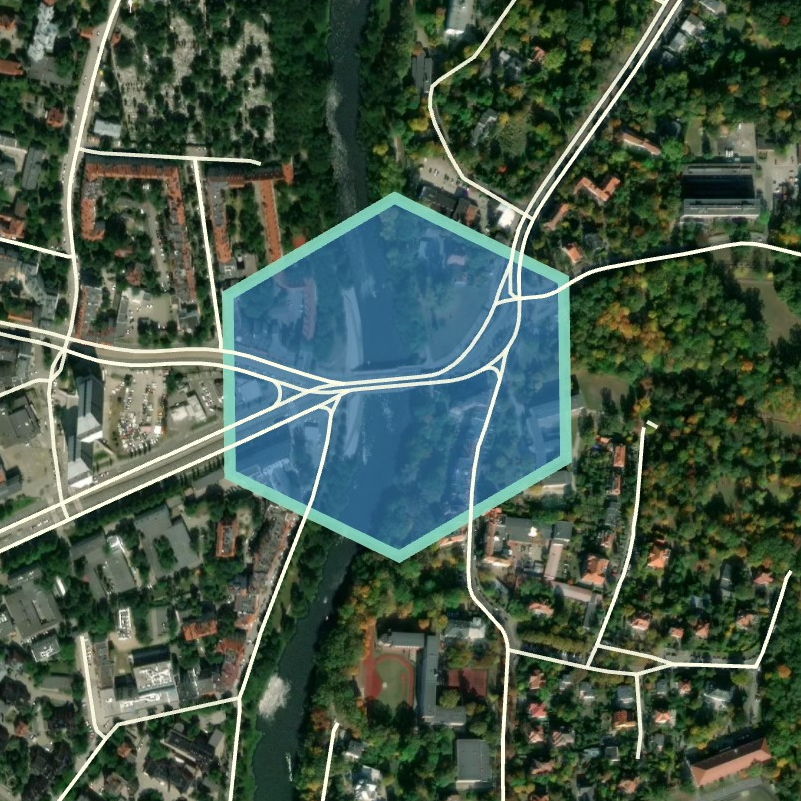}
                    \caption{High-traffic region containing a bridge in Wrocław, Poland.}
                \end{subfigure}
                \hfill
                \begin{subfigure}[t]{0.32\textwidth}
                    \centering
                    \includegraphics[width=\linewidth]{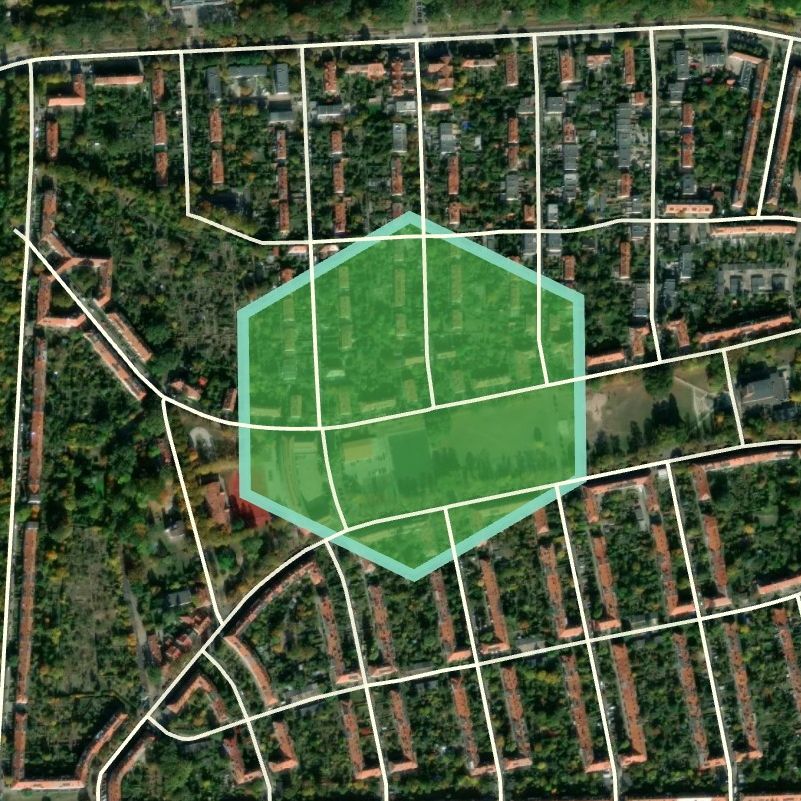}
                    \caption{Residential region in Wrocław, Poland.}
                \end{subfigure}
                \hfill
                \begin{subfigure}[t]{0.32\textwidth}
                    \centering
                    \includegraphics[width=\linewidth]{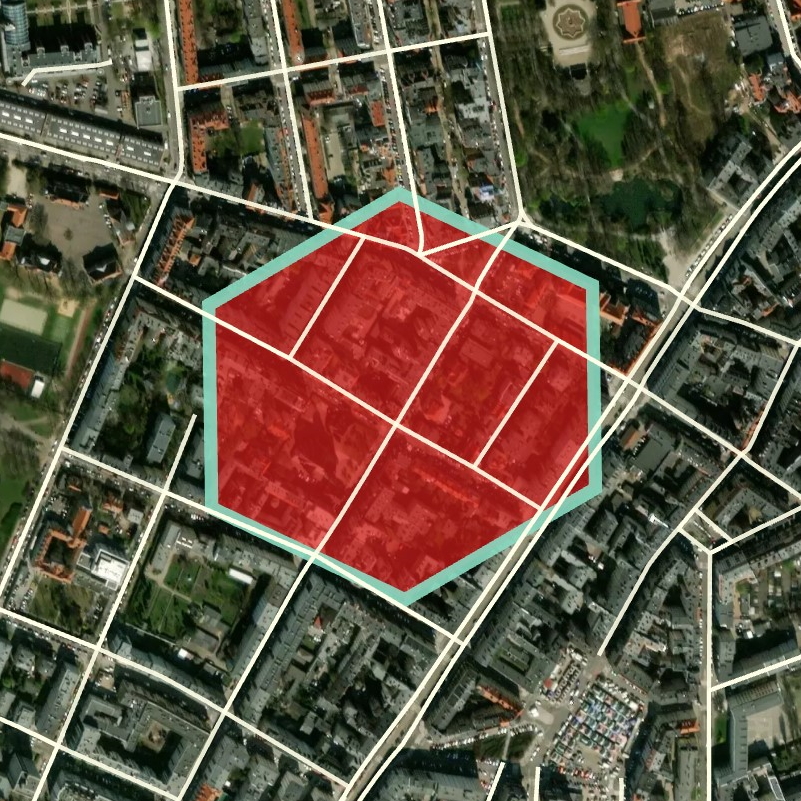}
                    \caption{Residential area next to high-traffic roads in Poznań, Poland.}
                \end{subfigure}
                
                \caption{Addition in the obtained embedding space (\textit{Blue} + \textit{Green} = \textit{Red})}                
                \label{fig:operations_addition}
            \end{figure}
            
            First, the \textbf{addition} of two vectors is performed, conditioned on the resulting embedding being in \textit{Poznań}. Two microregions from \textit{Wrocław} are selected, which are distinct in their nature and serve one purpose in the city road infrastructure. These two regions are high-traffic (with a bridge) and residential regions, respectively. The sum of the embeddings yields a functionally merged region, which is a residential region adjacent to high-traffic roads. This process is shown in Figure \ref{fig:operations_addition}. Looking at the feature profiles of the microregions, one can see that share-wise the resulting zone is an imperfect merger of the summands. However, the most important tag that is not taken into account is \textit{bridge}, which is at the core of the second microregion.
        
\section{Conclusions}
\label{ch:conclusion}
We propose a method for the unsupervised embedding of microregions using road infrastructure information and use the obtained latent space to conduct similarity detection, and as a result, define a typology of city regions. To accomplish the stated goals, several subtasks were specified and completed.
      
Using the proposed method, the road segments for $13$ Polish cities were embedded, which later were aggregated to their corresponding microregions, eventually creating their representations. Then, similarity detection was performed on the obtained latent space using \textit{Agglomerative Clustering}. This resulted in the grouping of the different numbers of clusters. Each such division was later discussed and analyzed, which in the end resulted in the high-level definition of region typology based on road infrastructure information. Finally, the obtained latent space was analyzed, and arithmetic operations such as addition and subtraction of the embedding vectors were performed. Carrying out the experiments proved that the obtained representations carry meaningful information and the semantic meaning of the real-world objects is preserved.
    
Although the proposed solution seems simple, the final results prove that it could capture the characteristics of the road network. Furthermore, it provided diversified enough urban zone representations upon which the region typology could be defined. Additionally, an automatic way of discovering the city's road structure was identified.

\section{Future works}
    We believe that the proposed methodology provides the backbone for further development in the area of geospatial data, and more specifically, in the region embedding using road infrastructure information. Based on the input of this thesis, some further improvements can be identified, such as:
    \begin{itemize}
        \item graph aspect - road infrastructure is intrinsical of graph nature. One could use methods that use this type of data fully. Previous works experimented with this approach but lacked in using spatial indexes,
        \item roads position and orientation - this thesis does not consider these aspects. Enriching the model with such information could further improve the model,
        \item model change - the \textit{autoencoder} used in this thesis is a basic approach. One can try utilizing more advanced solutions - \textit{VAE} \cite{vae} or \textit{beta-VAE} \cite{b-vae} - or try \textit{Skip-Gram} \cite{skip-gram} approach to geospatial data, as used in \cite{hex2vec},
        \item downstream tasks - evaluating method quantitatively on predictive tasks would enable comparing the obtained embeddings to other methods,
        \item intersections - data available in the intersections could be incorporated into the method,
        \item network types - this thesis was based on driveable road networks. One could try to apply the methodology to other types of road networks (bicycle road infrastructure, railways, etc.),
        \item data augmentation - publicly available \textit{open-data} sources such as \textit{OpenStreetMap} are not complete and a significant portion of the data is missing. One could try to impute the tags using other present features, and national or domestic laws,
        \item region aggregation - in this thesis, the road embeddings were equally aggregated into the corresponding regions. Weighting the road segments, e.g., by the road length in this region, could be beneficial,
        \item data range - incorporating more cities around the globe or even adding areas outside the cities could result in a more robust method.
    \end{itemize}

\bibliographystyle{ACM-Reference-Format}
\bibliography{sample-base}

\appendix

\section{Histograms and typology maps}

    \begin{figure*}[!h]
        \centering
        \begin{subfigure}[b]{0.49\textwidth}
            \centering
            \includegraphics[width=\linewidth]{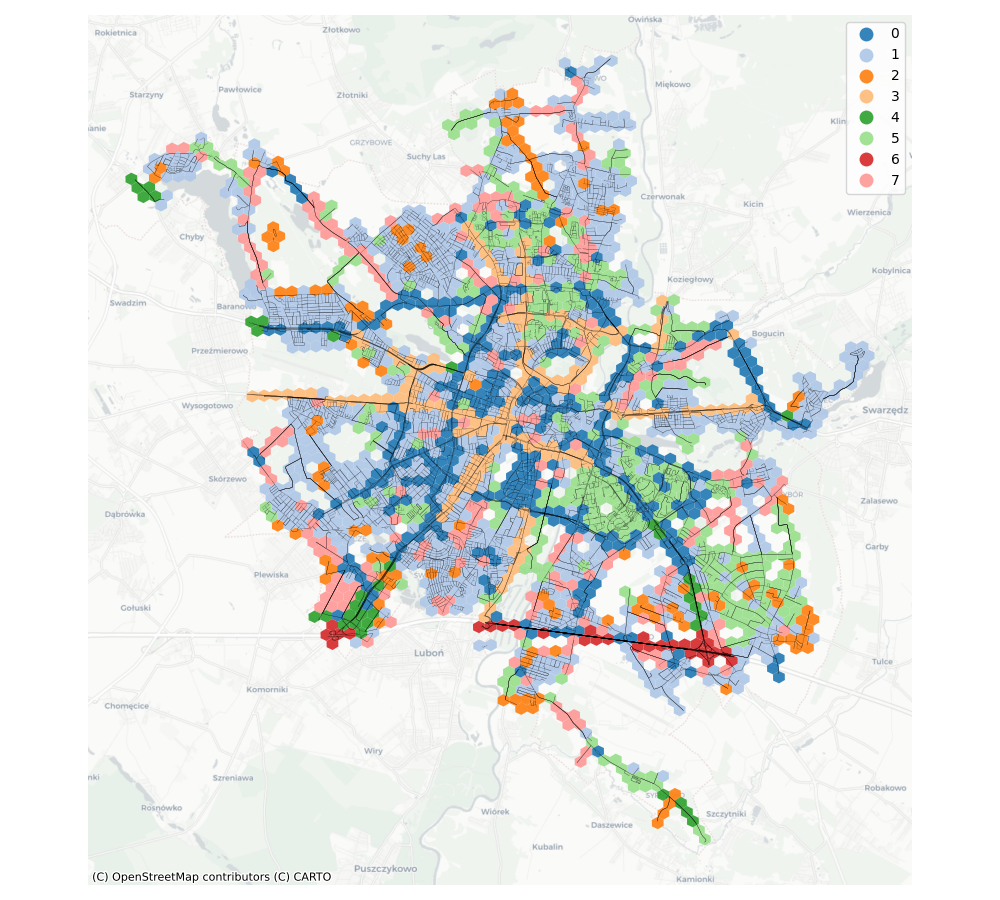}
            \caption{Poznań}
        \end{subfigure}
        \hfill
        \begin{subfigure}[b]{0.49\textwidth}
            \centering
            \includegraphics[width=\linewidth]{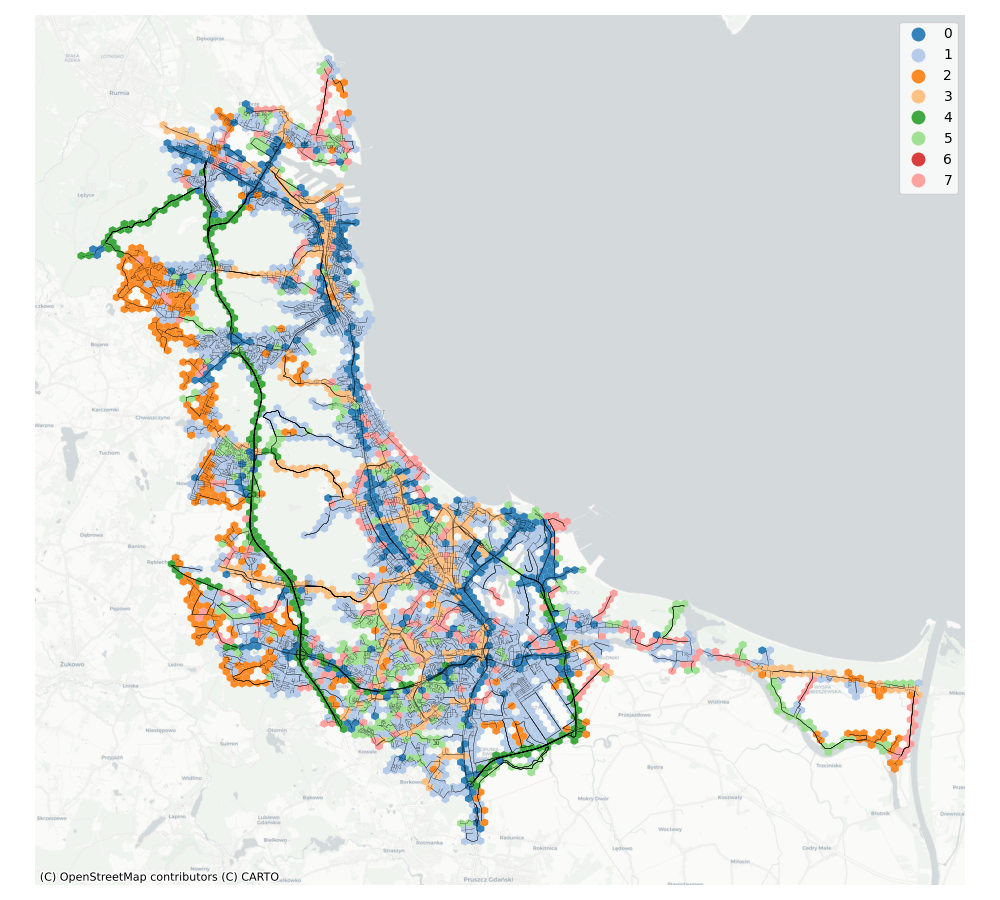}
            \caption{Trójmiasto (Gdańsk, Gdynia, Sopot)}
        \end{subfigure}
        
        \vskip 0pt
        
        \begin{subfigure}[b]{0.49\textwidth}
            \centering
            \includegraphics[width=\linewidth]{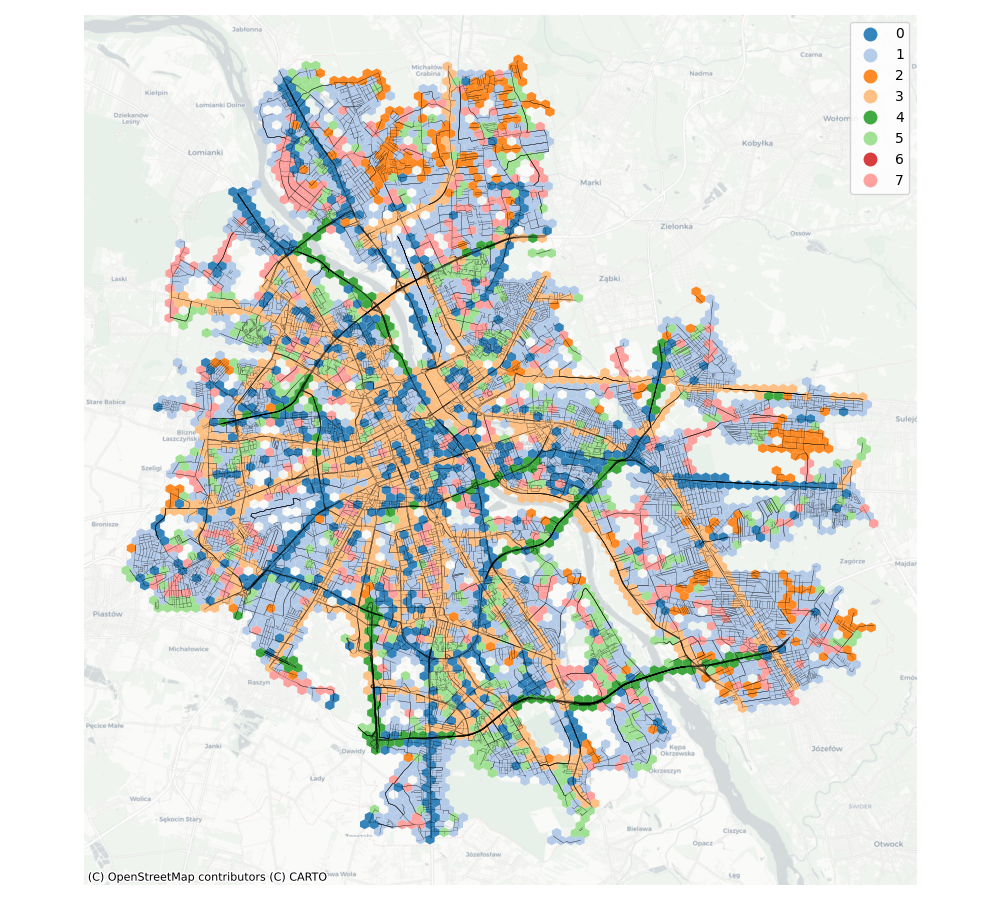}
            \caption{Warszawa}
        \end{subfigure}
        \hfill
        \begin{subfigure}[b]{0.49\textwidth}
            \centering
            \includegraphics[width=\linewidth]{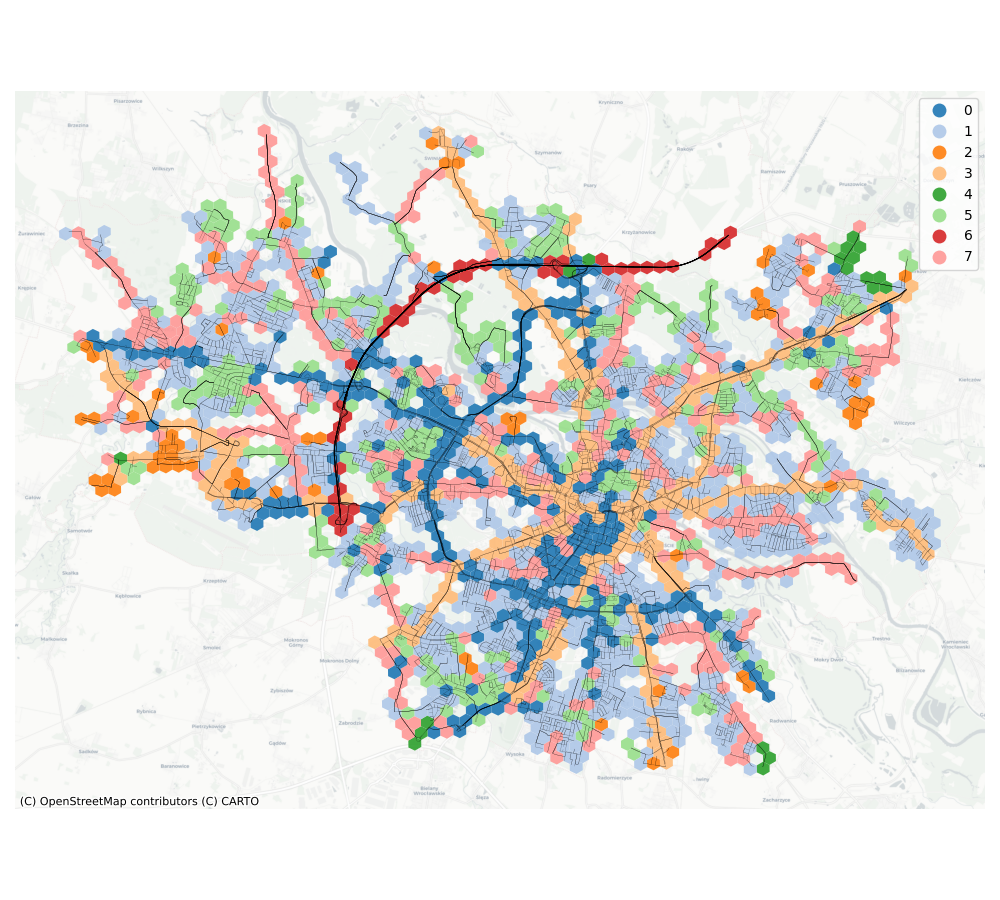}
            \caption{Wrocław}
        \end{subfigure}
        
        \caption{Microregion clustering results ($8$ clusters) for selected Polish cities}
        \label{fig:exp_hex_8}
    \end{figure*}

 \begin{figure*}[!htp]
    \centering
    
    \begin{subfigure}[b]{0.45\textwidth}
        \centering
        \includegraphics[width=\linewidth]{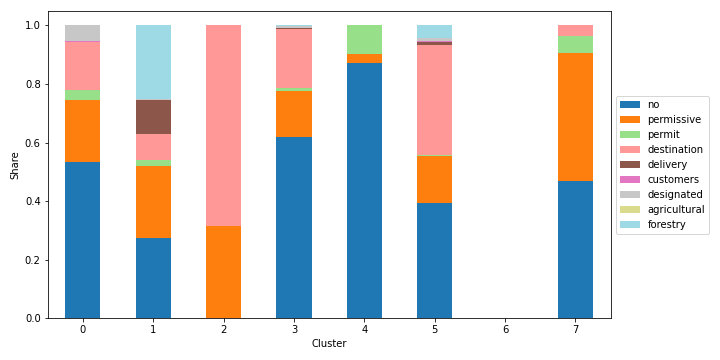}
        \caption{Access}
    \end{subfigure}
    \hfill
    \begin{subfigure}[b]{0.45\textwidth}
        \centering
        \includegraphics[width=\linewidth]{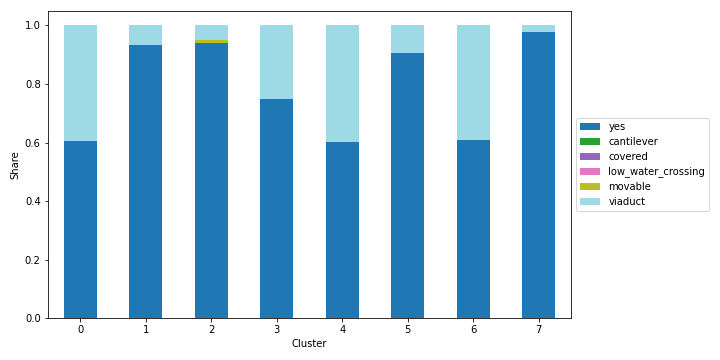}
        \caption{Bridge}
    \end{subfigure}
    \vskip 0pt
    \begin{subfigure}[b]{0.45\textwidth}
        \centering
        \includegraphics[width=\linewidth]{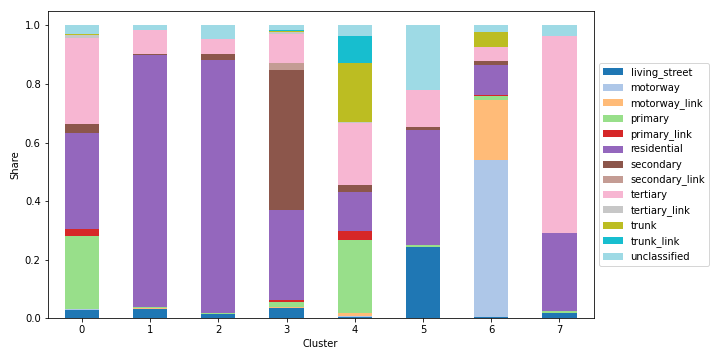}
        \caption{Highway}
    \end{subfigure}
    \hfill
    \begin{subfigure}[b]{0.45\textwidth}
        \centering
        \includegraphics[width=\linewidth]{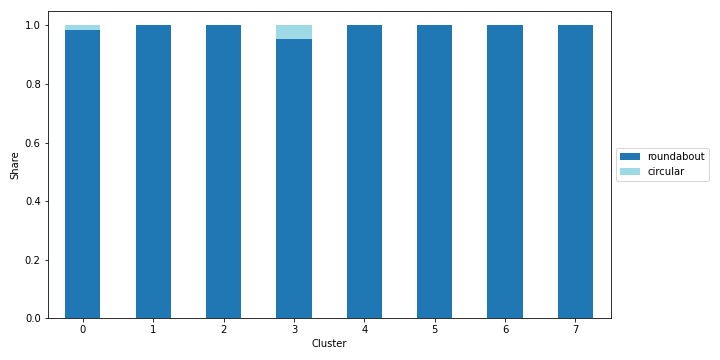}
        \caption{Junction}
    \end{subfigure}
    \vskip 0pt
    \begin{subfigure}[b]{0.45\textwidth}
        \centering
        \includegraphics[width=\linewidth]{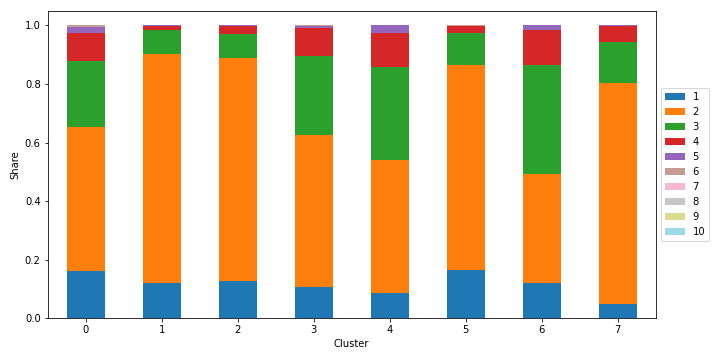}
        \caption{Lanes}
    \end{subfigure}
    \hfill
    \begin{subfigure}[b]{0.45\textwidth}
        \centering
        \includegraphics[width=\linewidth]{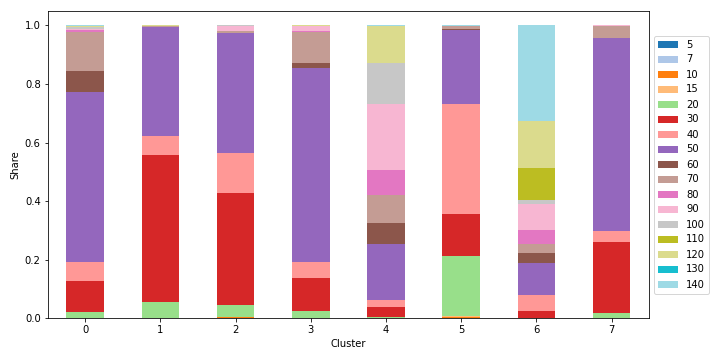}
        \caption{Maxspeed}
    \end{subfigure}
    \vskip 0pt
    \begin{subfigure}[b]{0.45\textwidth}
        \centering
        \includegraphics[width=\linewidth]{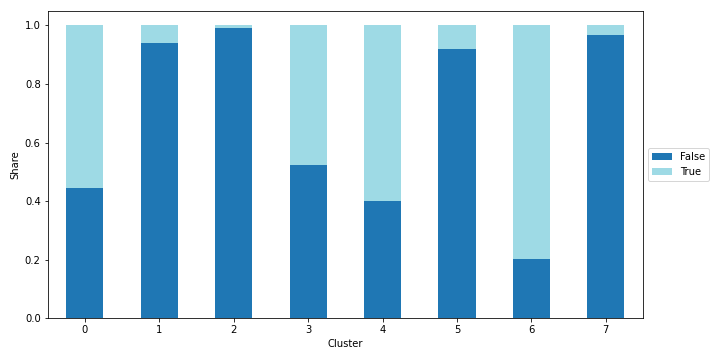}
        \caption{Oneway}
    \end{subfigure}
    \hfill
    \begin{subfigure}[b]{0.45\textwidth}
        \centering
        \includegraphics[width=\linewidth]{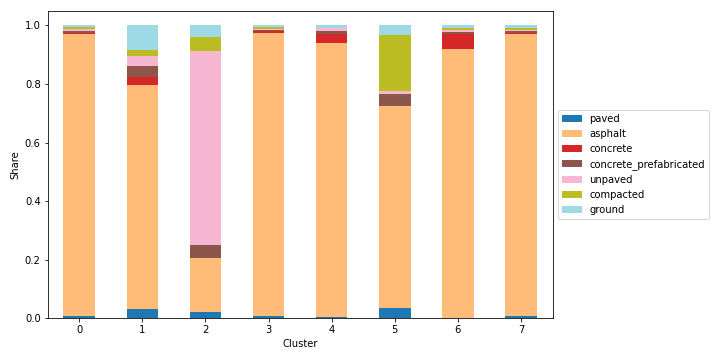}
        \caption{Surface}
    \end{subfigure}
    \vskip  0pt
    \begin{subfigure}[b]{0.45\textwidth}
        \centering
        \includegraphics[width=\linewidth]{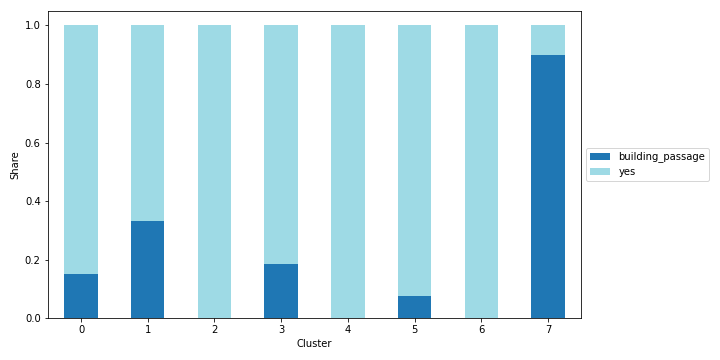}
        \caption{Tunnel}
    \end{subfigure}
    \hfill
    \begin{subfigure}[b]{0.45\textwidth}
        \centering
        \includegraphics[width=\linewidth]{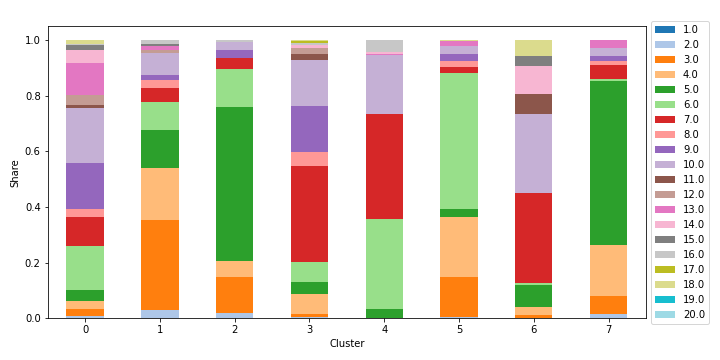}
        \caption{Width}
    \end{subfigure}
    
    \caption{Features' shares in final groups}
    \label{fig:features_clustering}
\end{figure*}

\end{document}